\documentclass{midl} 

\usepackage{booktabs}
\usepackage{multirow}
\usepackage{float}
\usepackage{tikz}
\usepackage{pgfplots}
\usepackage{array}

\pgfplotsset{compat=1.18}
\jmlrvolume{-- 036}
\jmlryear{2025}
\jmlrworkshop{Full Paper -- MIDL 2025}
\editors{Accepted for publication at MIDL 2025}

\title[Generate to Ground]{Generate to Ground: Multimodal Text Conditioning Boosts Phrase Grounding in Medical Vision-Language Models}

 \midlauthor{\Name{Felix Nützel\nametag{$^{1}$}} \Email{felix.nuetzel@fau.de} \\
             \Name{Mischa Dombrowski\nametag{$^{1}$}} \Email{mischa.dombrowski@fau.de} \\
             \Name{Bernhard Kainz\nametag{$^{1,2}$}} \Email{bernhard.kainz@fau.de} \\
             \addr $^{1}$ Friedrich-Alexander-Universität Erlangen-Nürnberg, DE \\
             \addr $^{2}$ Imperial College London, UK}

\begin{document}

\maketitle

\begin{abstract}
Phrase grounding, \emph{i.e.}, mapping natural language phrases to specific image regions, holds significant potential for disease localization in medical imaging through clinical reports. While current state-of-the-art methods rely on discriminative, self-supervised contrastive models, we demonstrate that generative text-to-image diffusion models, leveraging cross-attention maps, can achieve superior zero-shot phrase grounding performance. 
Contrary to prior assumptions, we show that fine-tuning diffusion models with a frozen, domain-specific language model, such as CXR-BERT, substantially outperforms domain-agnostic counterparts. This setup achieves remarkable improvements, with mIoU scores doubling those of current discriminative methods. These findings highlight the underexplored potential of generative models for phrase grounding tasks. 
To further enhance performance, we introduce Bimodal Bias Merging (BBM), a novel post-processing technique that aligns text and image biases to identify regions of high certainty. BBM refines cross-attention maps, achieving even greater localization accuracy. Our results establish generative approaches as a more effective paradigm for phrase grounding in the medical imaging domain, paving the way for more robust and interpretable applications in clinical practice. The source code and model weights are available at \url{https://github.com/Felix-012/generate_to_ground}.
\end{abstract}

\begin{keywords}
Phrase Grounding, Visual Grounding, Stable Diffusion, Latent Diffusion Models, Cross-Attention, Chest X-Rays
\end{keywords}

\section{Introduction}
\label{sec:introduction}
Phrase grounding refers to the ability of a model to map textual tokens to regions in an image.
Unlike typical object detection or segmentation tasks, phrase grounding usually takes natural language as input,
such as medical reports, instead of relying on a predefined set of categories.
Thus, phrase grounding can be seen as a generalization of object detection.

In the medical domain, phrase grounding can be used to localize anomalies in images based on textual descriptions
provided by experts~\cite{Bhalodia_2021}.
This is attractive, since it works without explicit labels, which are rare and expensive for medical data.
By inspecting phrase grounding performance, it is possible to infer which phrases or regions influenced the decision of the system,
assess whether the model made proper use of all available modalities, and if the modalities were aligned properly without
confusion~\cite{Parcalabescu_2020_phrase_grounding}.
These properties are vital for interpretability, which is a necessary requirement for models to be used in the medical field~\cite{Chen_2023}.
In addition, without interpretability, models could introduce harmful biases without explanations, which is
especially critical for high-risk decision-making~\cite{HAKKOUM2022_interpretability_medical}.

Existing discriminative medical phrase grounding approaches can be roughly categorized into two groups: 
supervised and self-supervised with contrastive
learning~\cite{Boecking_2022_MS_CXR,Gupta_2020,zhang_2022}.
An important supervised approach is MedRPG~\cite{Chen_2023}, which uses ground-truth bounding boxes of radiographs to formulate a contrastive loss based on the features of bounding boxes, as well as the joint attention of bounding boxes, the class token and an additional learnable token.
Another relevant branch of medical phrase grounding methods are those that work with 3D medical data, such as the paper by~\citet{Ichinose_2023}, which addresses the unique issues of phrase grounding in CT scans.
They suggest using a pre-trained segmentation model that labels the anatomic structures visible in the scan and introduce a module to structure the corresponding medical reports.
However, such supervised methods require ground-truth bounding boxes or annotators, which are difficult to obtain, especially in the medical domain.
Self-supervised methods do not require explicit labels but do not always lead to the desired result.
Discriminative methods would usually evaluate their phrase grounding performance by computing the cosine similarity between the
text embeddings and the corresponding image embeddings.
However, it has recently been shown that phrase grounding tasks can also be solved using generative models in an unsupervised context \cite{Dombrowski_2024}.
Specifically, text-to-image Latent Diffusion Models (LDMs) are useful, due to their use of cross-attention to
combine the two modalities, as well as their ability to produce high-quality images.
Text-to-image LDMs are trained to generate images from a dataset while receiving additional text conditioning from the
corresponding text inputs ~\cite{Dombrowski_2023_ICCV,vilouras2024zeroshotmedicalphrasegrounding}.
Instead of using cosine similarity, the phrase grounding capabilities of LDMs are easier to evaluate by using their cross-attention layers.
Earlier, \citet{Dombrowski_2024} showed that using a frozen text encoder improves the phrase
grounding capabilities of an LDM.

So far, the self-supervised approach by~\citet{Boecking_2022_MS_CXR} achieved the highest phrase grounding performance on Chest X-ray (CXR) data
by fine-tuning a Large Language Model (LLM) pre-trained on the biomedical domain on CXR reports.
The resulting LLM is known as CXR-BERT.
This model is jointly trained with an image encoder, in a framework called BioViL.
In this work, we leverage these and use CXR-BERT as a frozen text encoder that conditions
the U-Net in an LDM.
Consequently, we inject the learned embeddings of CXR reports from CXR-BERT,
while additionally fine-tuning the U-Net on corresponding CXR images.
CXR-BERT and the LDM support each other in a bidirectional manner:
the LDM, having a generative architecture, is able to leverage the full phrase grounding potential of the text embeddings compared to the simple CNN that is used in BioViL.
Additionally, the powerful text embeddings learned by CXR-BERT provide the necessary conditioning to the LDM that enables the model to learn a well-grounded multimodal representation.

As a result, the contributions of our work include the following:
\begin{itemize}
\parskip0pt
    \item We demonstrate that a multimodal text encoder with domain-specific knowledge can vastly improve phrase grounding capabilities of an LDM.
    \item We show that generative approaches can yield far better phrase grounding results than traditional discriminative approaches by nearly doubling conventual performance metrics such as mIoU.
    \item We discuss a novel post-processing method that can boost the phrase grounding capabilities of phrase grounding frameworks.
\end{itemize}
\begin{figure}[t]
    \centering
    \includegraphics[width=0.9\textwidth]{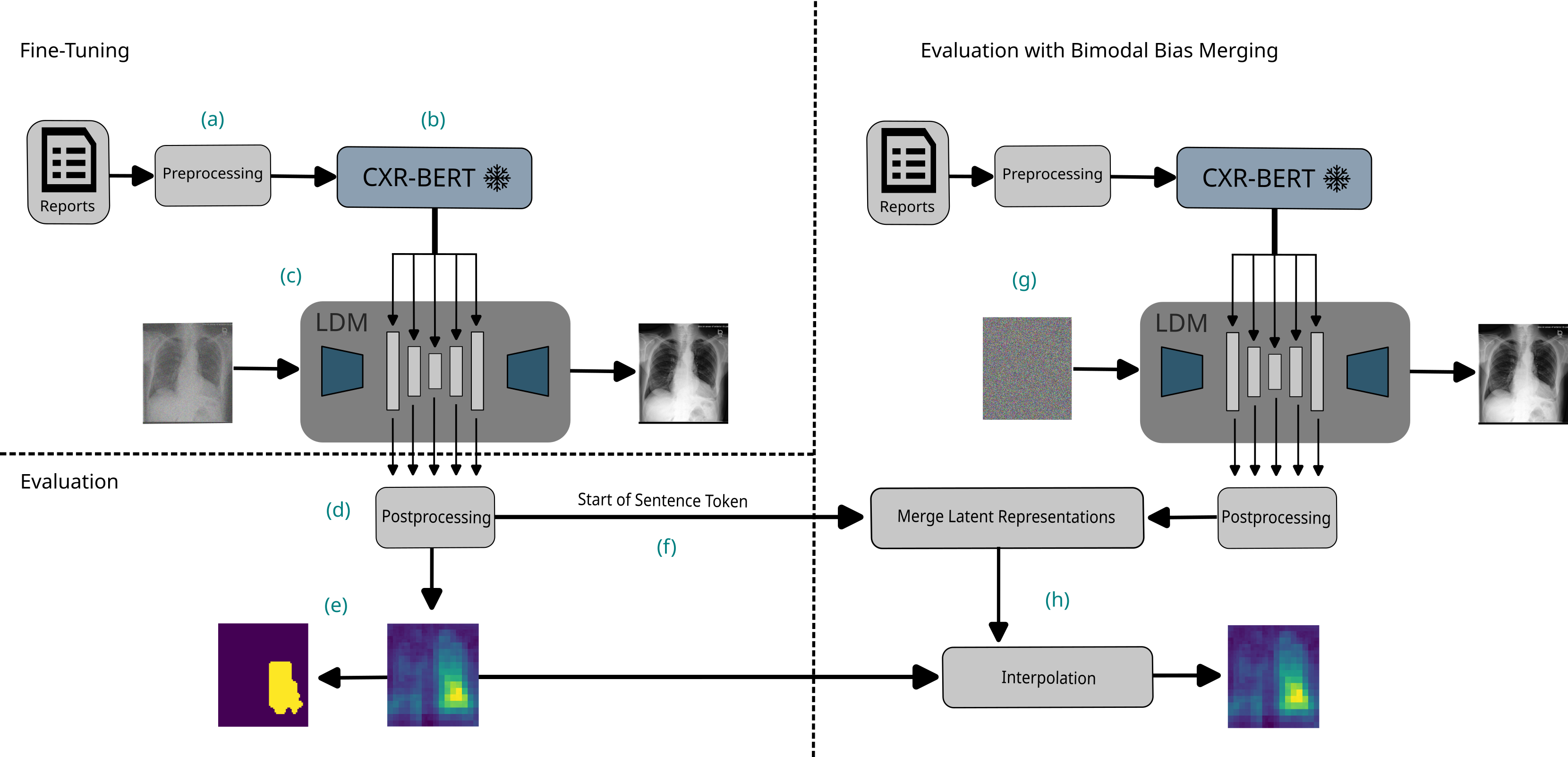}
    \caption{During fine-tuning, radiological text reports are extracted \textbf{(a)}.
    These reports are fed into CXR-BERT with frozen parameters \textbf{(b)}.
    The resulting text embeddings are used to condition the U-Net in the LDM, by injecting the embeddings into each cross-attention layer (represented as gray bars) in the U-Net.
    The LDM learns to generate images by getting noisy radiology images corresponding to the reports and a timestep as input \textbf{(c)}.
    During evaluation, noisy ground-truth images are repeatedly fed into the LDM to extract the corresponding cross-attention.
    These maps are processed based on relevant tokens and to get them into the correct dimensionality \textbf{(d)}.
    After a processing step, we obtain an activation map and its corresponding binary mask \textbf{(e)}.
    For BBM, we need to extract the image bias \textbf{(f)} and generate the text bias \textbf{(g)}, merge them and combine them with our original activation map \textbf{(h)}.}
    \label{fig:method}
\end{figure}
\section{Method}
\label{sec:method}
\subsection{Exchanging Text Encoders in Stable Diffusion}
\label{subsec:exchange}

Our approach is based on fine-tuning Stable Diffusion (SD)~\cite{Rombach_2022_CVPR}, a popular pre-trained text-to-image LDM.
We only train the underlying U-Net of SD while keeping the text encoder frozen, since~\citet{Dombrowski_2024}
demonstrated that maintaining the original configuration of the text encoder yields superior phrase grounding results.

We first fine-tune the U-Net by using the text encodings of the original pre-trained text encoder
(CLIP-ViT-L/14) of SD version 1.5~\cite{Rombach_2022_CVPR} on our training dataset.
Meanwhile, the parameters of the text encoder are kept frozen.
We then compare these baseline results by replacing the original text encoder with a frozen CXR-BERT~\cite{Boecking_2022_MS_CXR} encoder
in additional training runs.
CXR-BERT is a multimodal language model pre-trained on the CXR domain.
Due to being pre-trained on text and image inputs in a specific domain, CXR-BERT can provide
better text representations than the standard text encoder of SDs, which is trained on a largely domain-agnostic
subset of LAION-5B~\cite{Schuhmann_2022}.

Additionally, phrase grounding is inherently a task in which the image and the text modality need to be properly aligned.
Therefore, LLMs that received both vision and textual learning signals are especially suited for phrase grounding tasks.
This intuition is also supported by previous work, which already demonstrated that domain-specific text encoders with
multimodal pre-training perform well during phrase grounding tasks~\cite{Boecking_2022_MS_CXR}.
Still, some research, such as the paper by~\citet{Bluethgen_2024} suggests that using a domain-specific text encoder for LDMs does not yield any benefits in the CXR domain.
However, another interesting property of CXR-BERT~\cite{Boecking_2022_MS_CXR}, is the use of both global and local loss.
Typically, methods tend to use some variant of a global loss when pre-training a model, which computes the loss on image and phrase level.
But including a local loss, that aligns words with image regions, better reflects the bottom-up structure of phrase grounding, since each individual token is associated with a region in the image.

\subsection{Cross-Attention Map Extraction}
\label{subsec:extraction}
An overview of the cross-attention map extraction process can be seen on the left side of~\figureref{fig:method}.
The text inputs for CXR-BERT first need to be tokenized by its corresponding tokenizer with maximal
token length $N_{\max}$ into tokens $\tau_1, \dots, \tau_{N_{\max}}$.
Since only words with lexical meanings can be mapped to image regions, as demonstrated in~\figureref{fig:tokens}, we remove tokens corresponding to function words
by employing ScispaCy~\cite{neumann-etal-2019-scispacy}.
This approach yields a small improvement compared to the token processing method used by \citet{Dombrowski_2024} (see Appx.~\ref{sec:processing}).
From here, we are following the approach by~\citet{Dombrowski_2024}, meaning we are mostly interested in the probability matrix $P$, defined as
\begin{equation}
    \label{eq:attention}
     P = \text{softmax}\left(\frac{QK^T}{\sqrt{d_k}}\right),
\end{equation}
whereas $Q$ is the query, $K$ is the key and $d_k$ is the dimension of the attention embedding.
In particular, for the batch size $B$, the layers $L$, the image height $H$ and the image width $W$, $K$ is a learned linear projection of the text embeddings with dimension $(B * L \times N_{\max} \times d_k)$, while $Q$ is a learned linear projection of the image embeddings with dimension $(B * L \times H * W \times d_k)$.
Their inner product, the matrix $P$, is the basis for the cross-attention masks and is of dimension $(B * L \times H*W \times N_{\max})$.
Consequently, for each sample in the batch and for each layer, P sets each pixel and each token embedding in relation to each other, making it suitable to  evaluate visual grounding.
This matrix is generated and saved for each timestep during inference.
Therefore, when reshaping $P$ correctly and upsampling the image dimensions to our latent image size of 64, this results in a tensor with dimension
$(B \times T \times L \times N_{\max} \times 64 \times 64)$ for the number of timesteps $T$.
This allows us to easily select specific layers, timesteps and tokens.
The 2D activation maps $P_{comb}$ can then be obtained by simply averaging over the first three dimensions of each item in the
batch and excluding the start and end tokens.
This corresponds to computing the average over the attention maps for each timestep, layer and token.
When we use lexical filtering, only relevant tokens are used for this averaging.
An intuition for this approach is provided by~\figureref{fig:tokens}.
Unlike other text encoders that would result in focusing on unnecessary details such as the ribcage for attention maps in higher resolutions, CXR-BERT produces strong attention maps across all resolutions in the U-Net, which is why we can average over the layers (see Appx.~\ref{sec:layers} for an example).
Due to this consistency of CXR-BERT, we also do not need to select specific timesteps.
To obtain accurate localization capabilities from these maps, the model needs both an image input and a textual input.
Therefore, instead of starting with Gaussian noise during sampling, the ground-truth image is used as input in each
timestep.
The appropriate noise for the current timestep is added to a fresh input image in each step.
The corresponding binary mask for computing mIoU is obtained via fitting a Gaussian Mixture Model to the activation map.

\subsection{Bimodal Bias Merging}
\label{subsec:bias}
To further improve the accuracy of the 2D activation maps $P_{comb}$, we incorporate more information via a process we call Bimodal Bias Merging (BBM).
An overview of this method can be seen on the right side of~\figureref{fig:method}.
In this process, we combine the activation maps from~\sectionref{subsec:extraction} with the textual bias and image bias of the model, as motivated by the results of~\sectionref{sec:phrase-grounding-benchmarks}.
To this end, we only extract the cross-attention values of $P$ that correspond to the start token, which
represents the image bias of the model.
This way, we end up with a tensor $P_\text{img}$ of dimension $(B \times T \times L \times 1 \times 64 \times 64)$.
For the textual bias of the model, we need to sample the LDM again, but with the usual Gaussian noise as image input this time.
This results in a tensor $P_\text{txt}$ of dimension $(B \times T \times L \times N_{\max} \times 64 \times 64)$.
By combining $P_\text{txt}$ and $P_\text{img}$ via matrix multiplication (denoted as $\otimes$), we capture cross-modal interactions between the two representations.
Empirically, $P_\text{mult} = P_\text{img} \otimes P_\text{txt}$ consists of large radial gradients that show the most likely locations of the disease.
To have a measure for the accuracy of this map, we compute the structural similarity index measure~\cite{ssim_wang} $s$ between the
map for the text bias and the image bias, clipped to the range $[0,1]$.
Finally, to obtain our new activation map $P_\text{BBM}$, the bias interaction map and the original activation map are interpolated via the following quadratic Bézier curve:
\begin{equation}
    \label{eq:lbm}
    P_\text{BBM} = 2(1-s)s\left(\frac{P_\text{mult} + P_{\text{comb}} + P_\text{mult} \odot P_{\text{comb}}}{2}\right) 
+ (1-s)^2 P_{\text{comb}}
+ s^2 P_\text{mult}
\end{equation}
with $\odot$ being the Hadamard product.
This interpolation is essentially linear, except for the control point receiving additional information regarding the multiplicative interaction between the biases.
In its base form, BBM is the linear interpolation $sP_{\text{comb}} + (1-s)P_{\text{mult}}$, which improves the activations around the location of the disease, as can be measured with CNR.
However, this typically does not improve the generated attention maps, as measured with mIoU, since thresholding would include even low activations as part of the masks.
For this purpose, we introduced a control point to the equation that essentially serves as a gating mechanism that constricts the activation areas to adhere to the merged biases. 
Therefore, it is primarily relevant when computing masks.
By construction, Equation~\ref{eq:lbm} remains close to a linear interpolation, despite utilizing the gating mechanism, which considerably improves the activation maps, while also giving a slight boost to the masks.
Since the activation maps are primarily supposed to increase interpretability, it should be easy to inspect them with the human eye.
The main benefit of using BBM is that the activations are much clearer to see, which is more meaningful than simply using masks.
For more details on the interpolation, see Appx.~\ref{sec:processing}.

In this way, the original map is combined with the modality interaction map based on the calculated confidence score.
This is based on the heuristic that, if the image bias and text bias are similar, then the fields created by their merging are more likely to support finding an accurate location of the disease.
Meanwhile, if the two biases have large discrepancies, their combined information is less likely to enhance the original activation map and should mostly be ignored.

\section{Experiment Setup}
\label{sec:experiment_setup}
\subsection{Training Setup}
\label{subsec:training-&-optimization-setups}
Each run is performed using the same configuration:
we use a base learning rate of $5e^{-05}$, with 1000 warmup steps and a cosine learning rate scheduler.
The training is split over eight 80GB A100 GPUs with a batch size of 16 and two gradient accumulation steps each,
resulting in an effective batch size of 256.
To be reproducible, the used seeds were uniformly sampled and are 4200, 1759 and 6357.
For more efficiency, the model weights are converted to mixed precision.
During training, unconditional guidance training is applied, so the text conditioning would be dropped with a probability of 30\%.
Additionally, we keep an Exponential Moving Average of our U-Net, which is used for sampling.

\subsection{Datasets}
\label{subsec:datasets}
The base dataset for training and testing is the MIMIC-CXR dataset~\cite{Goldberger_2000_PhysioNet,Johnson_2019_MIMIC}. 
It contains pairings of CXR images and their respective reports, which include medical findings such as diseases. 
For training, we use the train split proposed by MIMIC-CXR-JPG~\cite{Goldberger_2000_PhysioNet,Johnson_2024_MIMIC_CXR_JPG}, which consists of 162,651 image-report pairs.
Explorative hyperparamter optimizations were conducted on the ChestXRay14 dataset~\cite{wang_2017_chestxray}.

Our test set consists of MS-CXR~\cite{Boecking_2022_MS_CXR}, a subset of MIMIC-CXR.
MS-CXR features improved bounding boxes, which can be used to evaluate the phrase-grounding performance of our models. 
Additionally, MS-CXR includes refined report descriptions that yield higher evaluation accuracy.

\subsection{Metrics}
\label{subsec:metrics}
In order to be comparable with~\citet{Boecking_2022_MS_CXR}, we report the Contrast-to-Noise Ratio (CNR) and mean Intersection over Union (mIoU) of our results.
CNR is calculated as $\text{CNR} = \frac{|\mu_{A_i} - \mu_{A_e}|}{\sqrt{\sigma^2_{A_i} + \sigma^2_{A_e}}}$, where $\mu_{A_i}$ and $\mu_{A_e}$ represent the means and $\sigma^2_{A_i}$, $\sigma^2_{A_e}$ the variances of the similarity scores inside and outside the bounding box respectively. Therefore, CNR can be used to evaluate the phrase grounding performance without the need of applying a threshold~\cite{Boecking_2022_MS_CXR}.
Additionally, we compute mIoU as the mean of the Jaccard distances $J(A, B) = \frac{|A \cap B|}{|A \cup B|}$ between the overlapping and non-overlapping regions of the thresholded phrase grounding image and the ground-truth bounding box.

\section{Results \& Discussion}
\begin{figure}[t]
    \centering
    \resizebox{0.85\textwidth}{!}{%
    \begin{tabular}{c c c c c c c c}
        \includegraphics[width=0.11\textwidth, height=0.11\textwidth]{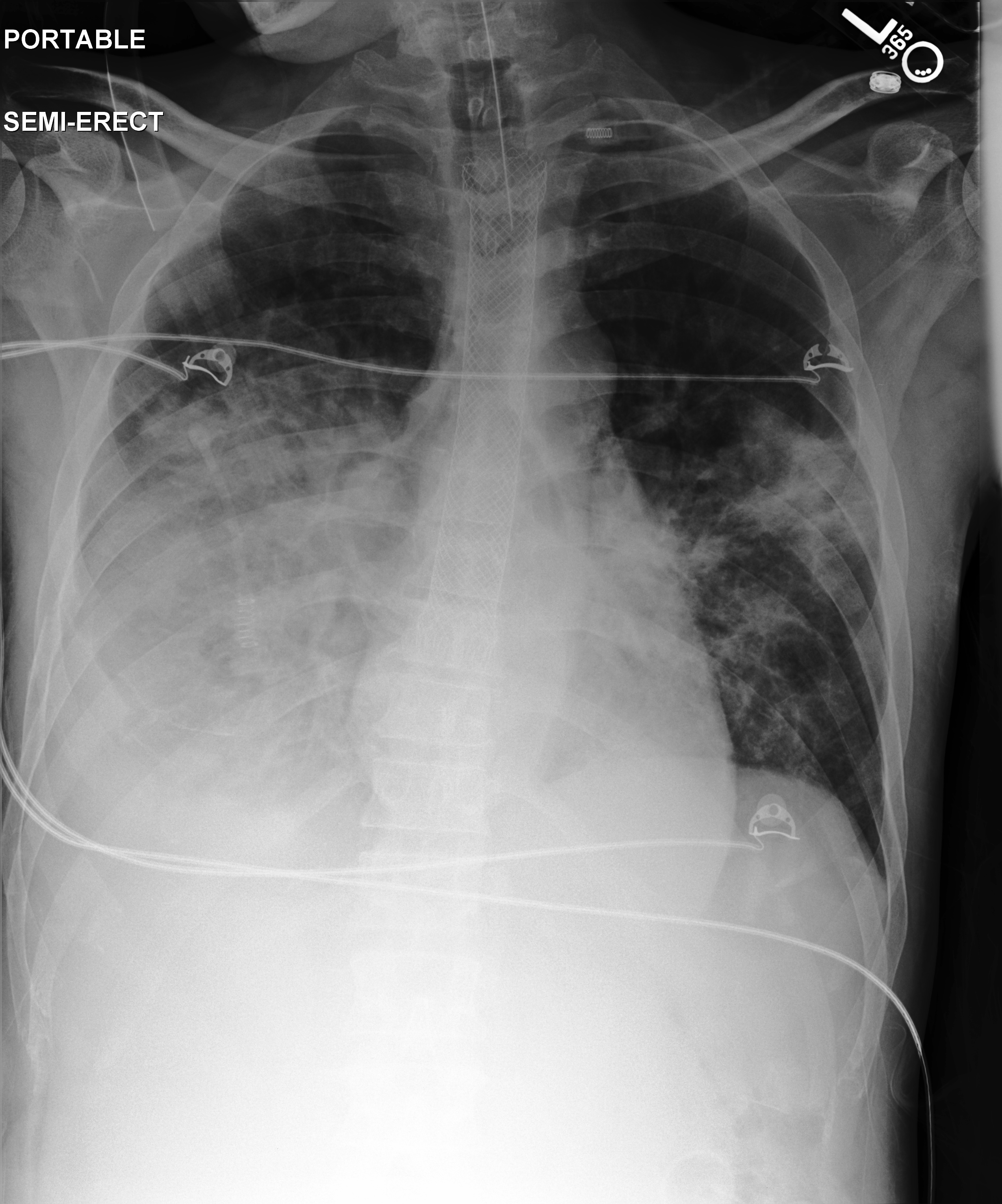} &
        \includegraphics[width=0.11\textwidth]{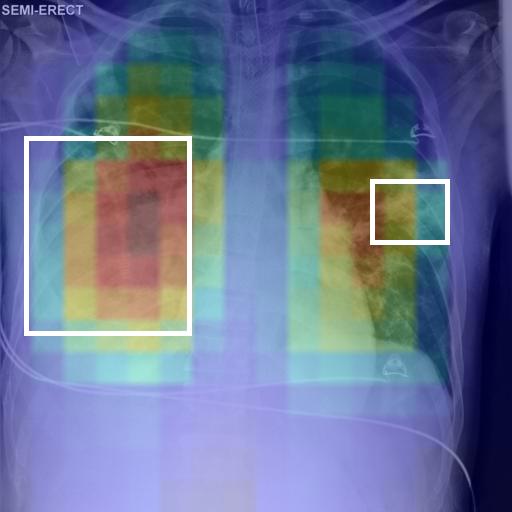} &
        \includegraphics[width=0.11\textwidth]{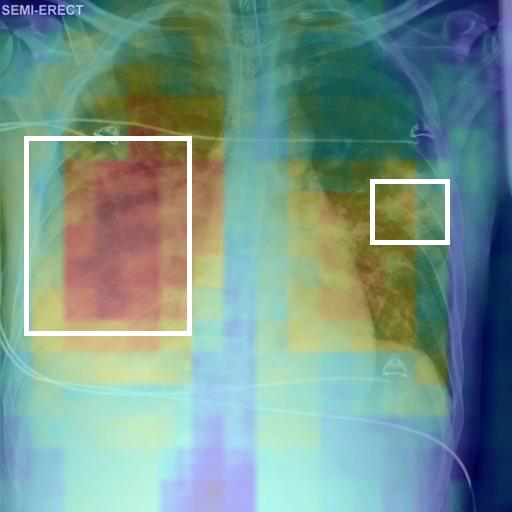} &
        \includegraphics[width=0.11\textwidth]{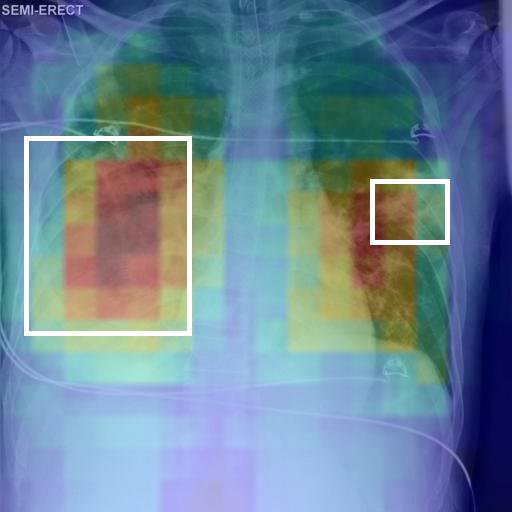} &
        \includegraphics[width=0.11\textwidth]{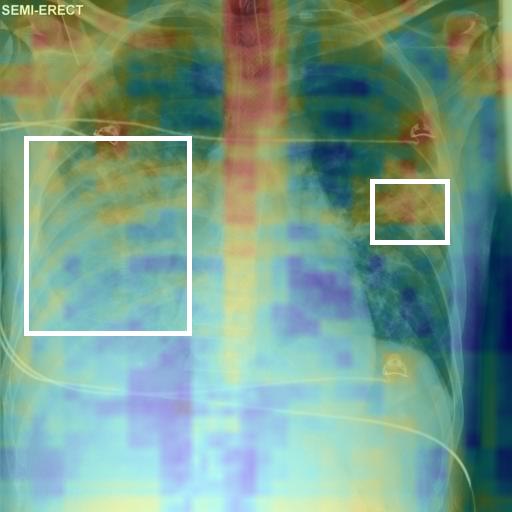} &
        \includegraphics[width=0.11\textwidth]{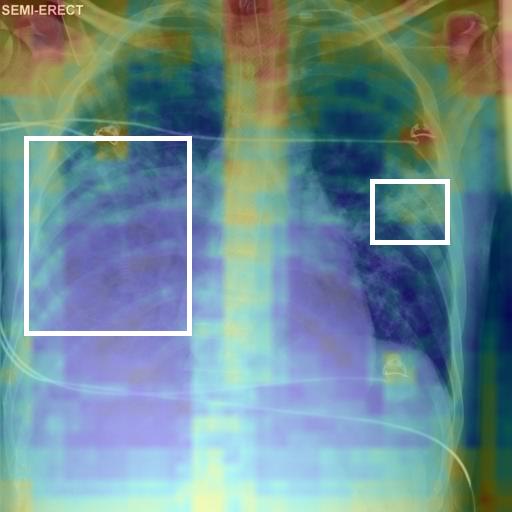} &
        \includegraphics[width=0.11\textwidth]{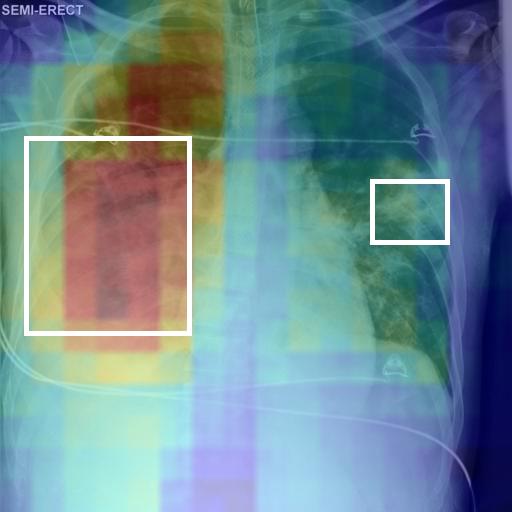} &
        \includegraphics[width=0.11\textwidth]{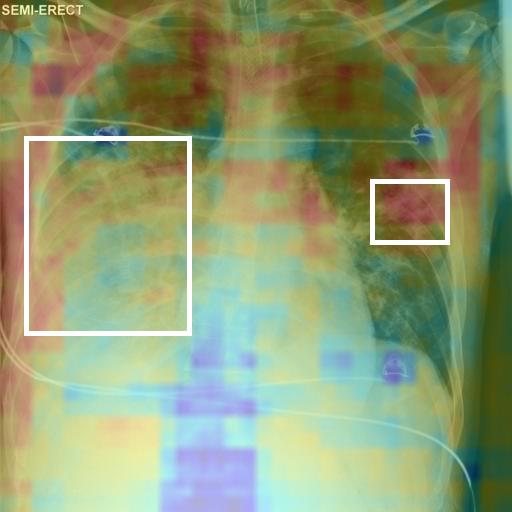} \\
        \textit{} & \textit{[start]} & \textit{patchy} & 
        \textit{consolid.} & \textit{in} & \textit{the} & \textit{right} & \textit{mid} \\
        \rule{0pt}{0.2cm} \\
        
        \includegraphics[width=0.11\textwidth]{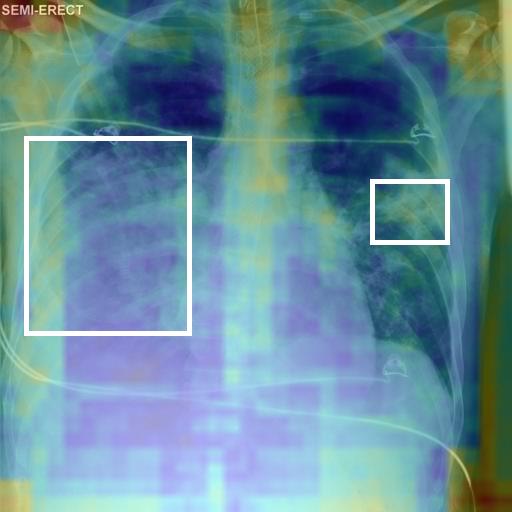} &
        \includegraphics[width=0.11\textwidth]{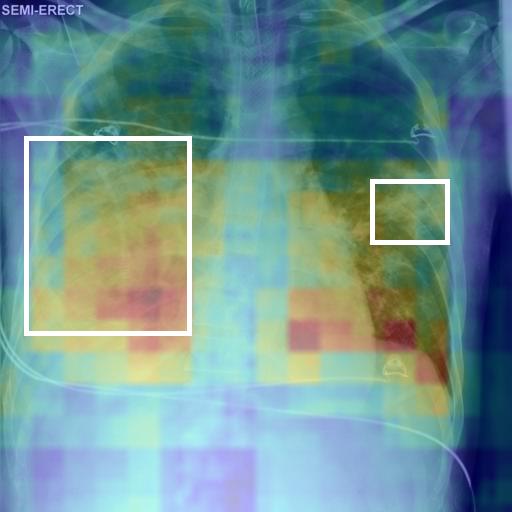} &
        \includegraphics[width=0.11\textwidth]{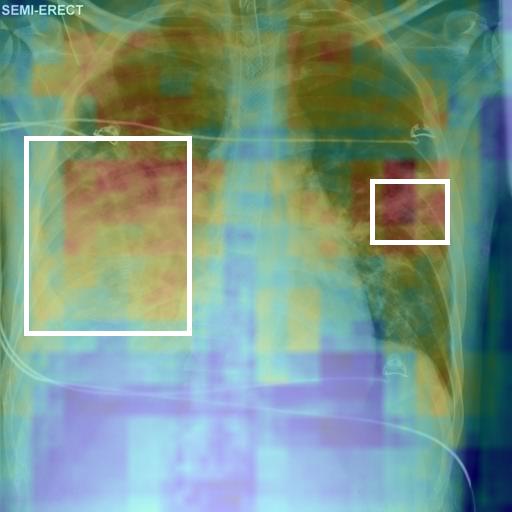} &
        \includegraphics[width=0.11\textwidth]{images/tokens/and} &
        \includegraphics[width=0.11\textwidth]{images/tokens/mid} &
        \includegraphics[width=0.11\textwidth]{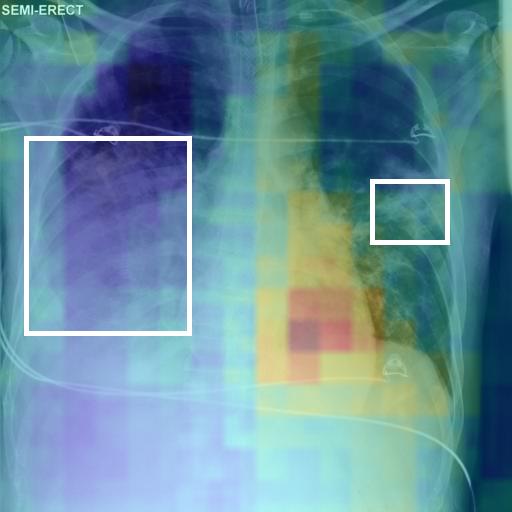} &
        \includegraphics[width=0.11\textwidth]{images/tokens/lung} &
        \includegraphics[width=0.11\textwidth]{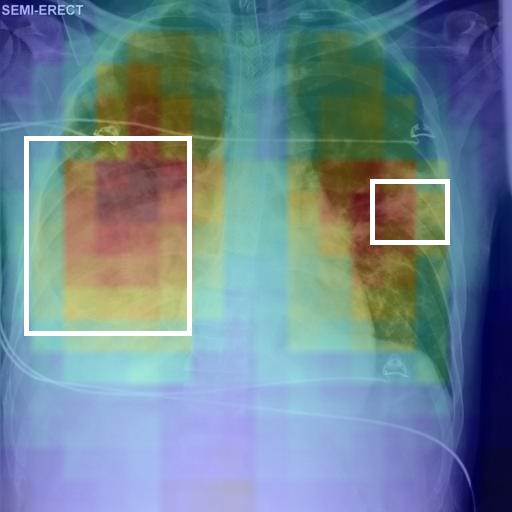} \\
        \textit{and} & \textit{lower} & \textit{lung} & \textit{and} & \textit{mid} & \textit{left} & \textit{lung} & \textit{[end]} \\

    \end{tabular}
    }
    \caption{Shows an input CXR image in posterior-anterior view (so note that left and right are mirrored) together with the cross-attention maps for the tokens of the corresponding text report. Red shows high activations, while blue shows low activations. The white boxes indicate the ground-truth bounding boxes.}
    \label{fig:tokens}
\end{figure}

\label{sec:phrase-grounding-benchmarks}
In~\figureref{fig:tokens}, we can see how the model attends to different tokens in a sentence.
For example, the token for \textit{``consolidation''} shows the approximate region of the anomaly.
The activations for \textit{``right''} are on the right side, the one for \textit{``lower''} is on the lower part and so
on.
Meanwhile, tokens with no lexical content, such as \textit{``the''}, have no clear activation patterns.
The activation map of the start token is the image bias of the model, which works already remarkably well.
This implies that the model has a good internal representation of the diseases, even when considering the text and image modalities separately.
Meanwhile, the end token provides even better phrase grounding capabilities, since it can incorporate the knowledge of the preceding tokens.
Empirically, using the activation map of the end token produces similar results as using the mean of preceding tokens.

\begin{table}[t]
\label{tab:main_results}
\centering
\caption{Comparison of mIoU and CNR metrics on the MS-CXR dataset between CXR-BERT used as a text
encoder for a LDM (CXR-BERT\textsubscript{LDM}), in discriminative context (BioVil,
    adopted from~\citet{Boecking_2022_MS_CXR}, and with additional local loss (BioVil-L, adopted from ~\citet{Boecking_2022_MS_CXR}).
    Also includes results for our baseline using a frozen CLIP encoder (CLIP\textsubscript{LDM},
    reproduced from~\citet{Dombrowski_2024}.
    Our results (CXR-BERT\textsubscript{LDM}) are shown with BBM applied and without it, averaged over three training runs with different seeds (shown as mean$\pm$standard deviation).
    To be consistent with CLIP\textsubscript{LDM}, we also used 50 timesteps during inference, although our method requires far less (see Appx.~\ref{sec:timesteps}).}
\setlength{\tabcolsep}{6pt} 
\resizebox{\textwidth}{!}{%
\begin{tabular}{@{}lcccccccccc@{}}
\toprule
\multirow{2}{*}{Disease} & \multicolumn{2}{c}{BioVil} & \multicolumn{2}{c}{BioVil-L} & \multicolumn{2}{c}{CLIP\textsubscript{LDM}}
& \multicolumn{2}{c}{CXR-BERT\textsubscript{LDM}} & \multicolumn{2}{c}{+BBM} \\
\cmidrule(lr){2-3} \cmidrule(lr){4-5} \cmidrule(lr){6-7} \cmidrule(lr){8-9}  \cmidrule(lr){10-11}
                         & mIoU & CNR & mIoU & CNR & mIoU & CNR & mIoU & CNR & mIoU & CNR\\ \midrule
Atelectasis     & 0.296 & 1.02±.06 & 0.302 & 1.17±.04 & 0.425 & 1.14  & \textbf{0.58±.01} & 1.69±.02 & 0.55±.01 & \textbf{1.78±.01}\\
Cardiomegaly    & 0.292 & 0.63±.08 & 0.375 & 0.95±.21 & 0.451 & 0.75  & 0.65±.00 & 1.37±.01 & \textbf{0.66±.01} & \textbf{1.54±.04}\\
Consolidation    & 0.338 & 1.42±.02 & 0.346 & 1.45±.03 & 0.436 & 1.12 & \textbf{0.52±.00} & 1.62±.01 & 0.49±.00 & \textbf{1.68±.02}\\
Lung Opacity    & 0.202 & 1.05±.06 & 0.209 & 1.19±.05 & 0.402 & 1.20  & \textbf{0.46±.00} & 1.57±.03 & 0.45±.00 & \textbf{1.67±.03}\\
Edema           & 0.281 & 0.93±.03 & 0.275 & 0.96±.05 & 0.541 & 1.25  & \textbf{0.55±.01} & 1.35±.01 & 0.54±.01 & \textbf{1.37±.00}\\
Pneumonia       & 0.323 & 1.27±.04 & 0.315 & 1.19±.01 & 0.438 & 1.12  & \textbf{0.53±.00} & 1.60±.01 & 0.51±.01 & \textbf{1.71±.01}\\
Pneumothorax    & 0.109 & 0.48±.06 & 0.135 & \textbf{0.74±.05} &0.312 & 0.22 & \textbf{0.34±.01} & 0.45±.09  & 0.33±.01 & 0.28±.07\\
Pl. Effusion    & 0.290 & 1.40±.06 & 0.315 & 1.50±.03 & 0.356 & 0.73  & \textbf{0.55±.00} & \textbf{1.65±.04} & 0.51±.01 & 1.62±.05\\
\midrule
Weighted Avg.   & 0.266 & 1.03±.02 & 0.284 & 1.14±.04 & 0.409 & 0.72& \textbf{0.54±.00} & 1.21±.03 & 0.53±.01 & \textbf{1.26±.04}\\
\bottomrule
\end{tabular}
}
\end{table}

\begin{figure}[t]
    \centering
    \resizebox{0.6\textwidth}{!}{%
    \begin{tabular}{c c c c c}
        \textbf{Input} & \textbf{BioViL} & \textbf{CLIP\textsubscript{LDM}} & \textbf{CXR-BERT\textsubscript{LDM}} & \textbf{+BBM} \\
        \includegraphics[width=0.2\textwidth]{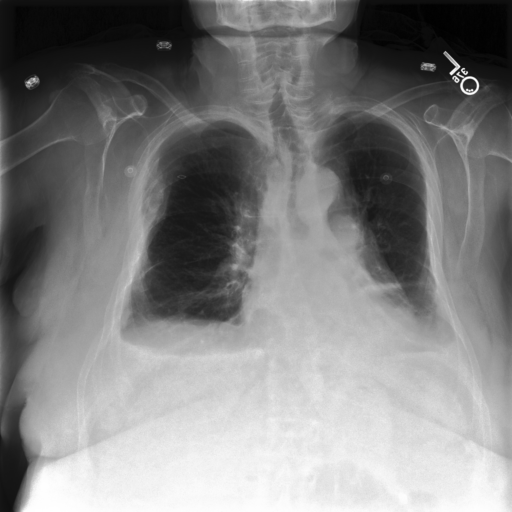} &
        \includegraphics[width=0.2\textwidth]{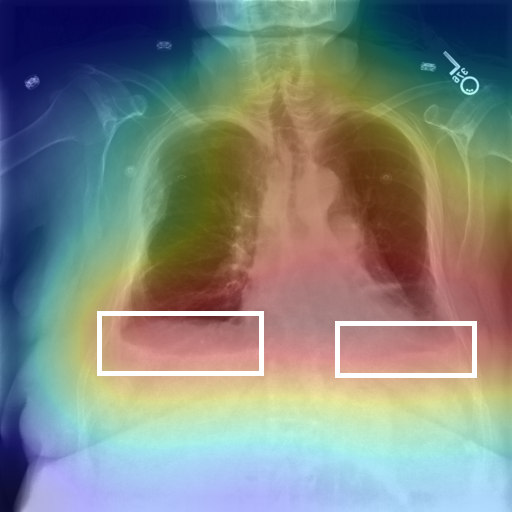} &
        \includegraphics[width=0.2\textwidth]{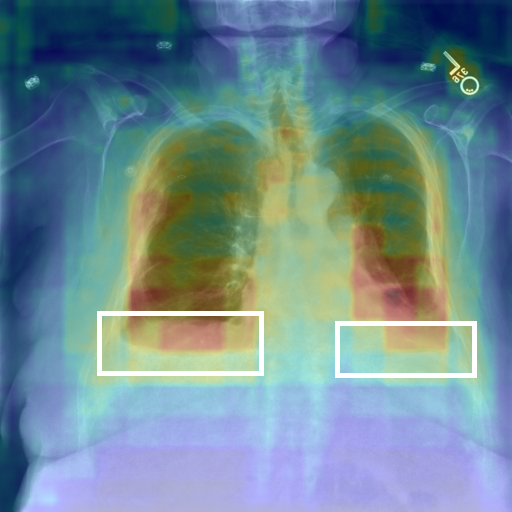} &
        \includegraphics[width=0.2\textwidth]{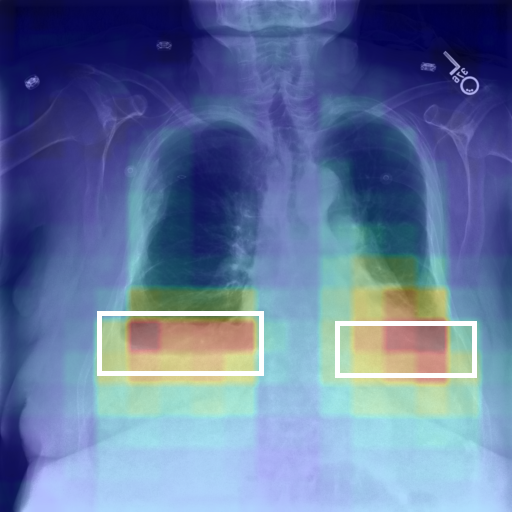} &
        \includegraphics[width=0.2\textwidth]{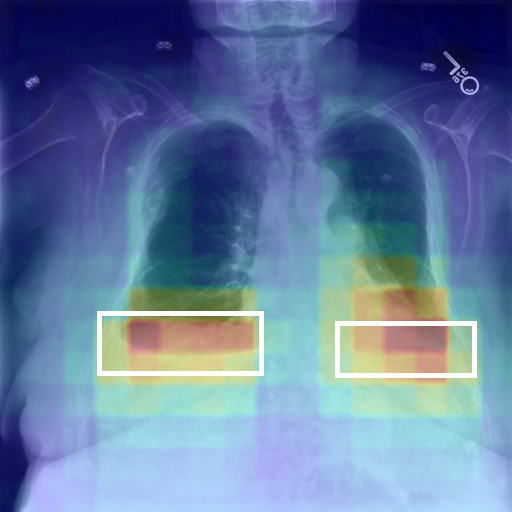} \\
        \textit{Pleural Effusion} & \textit{$CNR=1.46$} & \textit{$CNR=1.09$} & \textit{$CNR=2.32$} & \textit{$CNR=2.34$} \\
        \rule{0pt}{0.1cm} \\


        
        \includegraphics[width=0.2\textwidth]{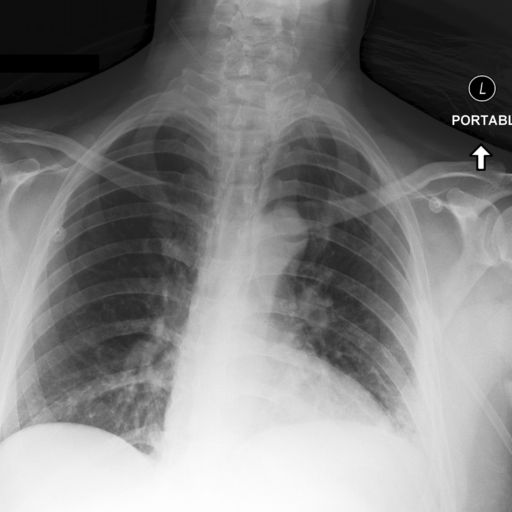} &
        \includegraphics[width=0.2\textwidth]{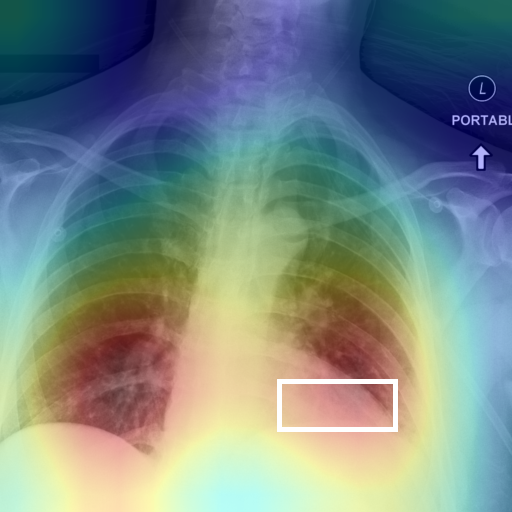} &
        \includegraphics[width=0.2\textwidth]{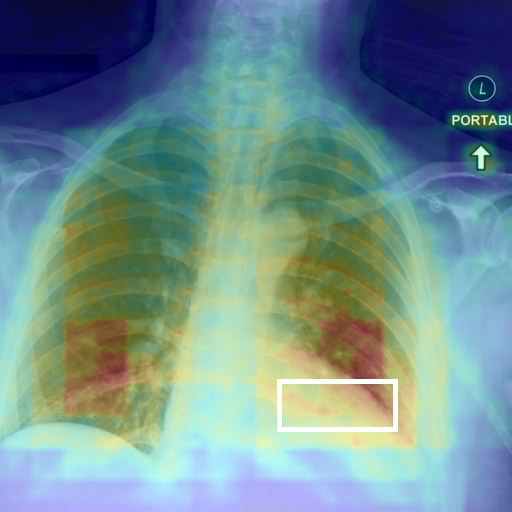} &
        \includegraphics[width=0.2\textwidth]{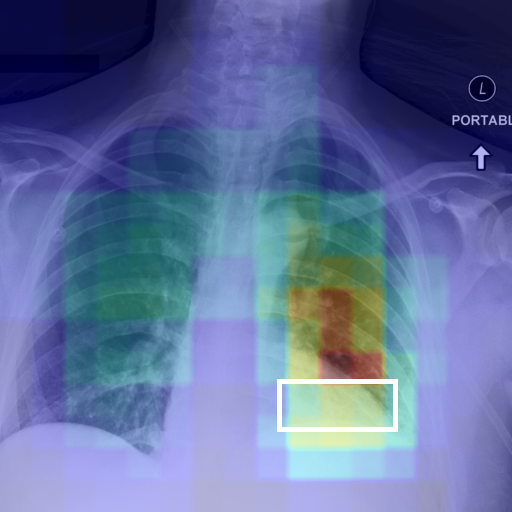} &
        \includegraphics[width=0.2\textwidth]{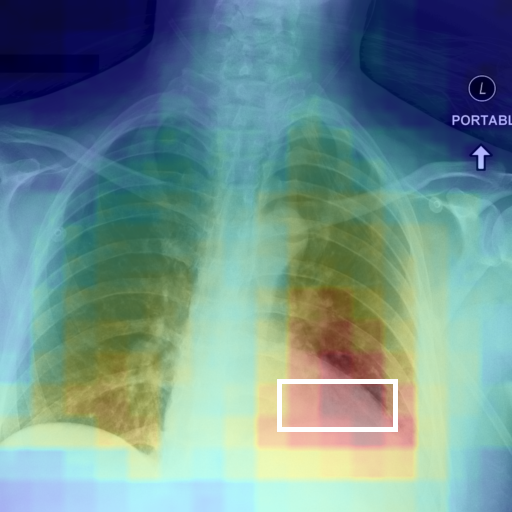} \\
        \textit{Lung Opacity} & \textit{$CNR=1.56$} & \textit{$CNR=1.41$} & \textit{$CNR=1.37$} & \textit{$CNR=2.24$} \\
    \end{tabular}
    }
    \caption{Examples for similarity maps of BioViL (results are reproduced from~\citet{Boecking_2022_MS_CXR}), as well as the cross-attention maps of a LDM with a frozen CLIP text encoder (CLIP\textsubscript{LDM}, reproduced from~\citet{Dombrowski_2024}), the LDM with a frozen CXR-BERT text encoder (CXR-BERT\textsubscript{LDM}) and the same encoder with additional BBM post-processing (+BBM). Red means higher activations, while blue means lower activations. The ground-truth bounding boxes are given in white. The respective input images with their disease labels are on the left.}
    \label{fig:examples}
\end{figure}

As shown in~\tableref{tab:main_results}, using CXR-BERT for text-conditioning in an LDM leads to
superior results in phrase grounding than using CXR-BERT in its original framework
BioViL~\cite{Boecking_2022_MS_CXR}.
The cross-attention maps of the LDM yield better results for both metrics (\textit{i.e.}, CNR and mIoU)
reported by~\cite{Boecking_2022_MS_CXR}.
Our approach approximately doubles the mIoU results.
Consequently, our approach seems to be especially suited for mask generation.
Additionally, our setup achieves higher results across all diseases compared to BioViL, and for almost all diseases
compared to its improved version, BioViL-L.
As discussed in Appx.~\ref{sec:tradeoff}, we can also confirm the observed trade-off between interpretability and image generation quality that was observed by~\citet{Dombrowski_2024}.
Our post-processing method BBM increases the CNR results considerably.
The improvement for mIoU is smaller, which most likely stems from the fact that for the mask generation, the applied thresholding destroys some of the gained information.
Notably, since our model could not learn pneumothorax properly, BBM decreases the phrase grounding performance for this disease.
These low values for pneumothorax align with the findings by~\citet{Dombrowski_2024}, suggesting that this disease
may be particularly challenging to model.
This might be due to Pneumothorax being very inconsistent in terms of location and size, or the corresponding impressions being too short.
Therefore, the model does not get enough information to model such a complex disease.

Furthermore, our approach improves the previous method by~\citet{Dombrowski_2024}, which is also based on the extraction of cross-attention maps.
However, the key difference is that we replaced the generic CLIP text encoder with the domain-specific
CXR-BERT text encoder.
This change greatly increases the performance of the model, as shown in~\tableref{tab:main_results}.
Consequently, we could show that a domain-specific LLM has the potential to greatly increase the phrase grounding potential of LDMs.
These results demonstrate that better phrase grounding can be achieved in a generative context compared to a
discriminative one, although the generative context has no specific alignment loss.

In~\figureref{fig:examples}, one can observe examples of cross-attention maps extracted from the LDM conditioned with a
CLIP text encoder and a CXR-BERT text encoder, as well as cosine similarity maps from BioViL.
As already demonstrated in~\citet{Dombrowski_2024}, employing a frozen CLIP text encoder yields solid phrase
grounding results.
However, it still performs worse than some domain-specific weakly supervised methods such as BioViL.
By employing text conditioning based on an encoder with strong phrase grounding capabilities in that domain,
the strengths of both methods are combined, resulting in the best outcomes.
As we can see in the second row of Figure~\ref{fig:examples}, BBM can sometimes correct inaccurate predictions made by our model, thus increasing its accuracy.

\noindent\textbf{Limitations:} 
First, while our model showed considerable improvements for the other diseases, it performed less optimally for Pneumothorax, which, as a disease, shows far less consistency in terms of shape and location in the body.
Due to these characteristics, Pneumothorax is a challenging disease to localize for both our method and comparing methods.
To properly learn to localize Pneumothorax, one example would be to directly fine-tune the model on a curated subset of the data, or alternatively, to take inspiration from other Pneumothorax detection works such as~\citet{Park_2022}, and apply knowledge distillation from a model fine-tuned specifically on Pneumothorax. 
However, we kept our fine-tuning process as general as possible to enable a fair comparison between methods.
Second, our BBM technique is strictly designed to complement models with strong phrase grounding capabilities and an LDM architecture. By focusing on well-aligned models, BBM optimally leverages their strengths, and the principles underlying our approach could inspire adaptations for other architectures or domains.
A detailed ablation study is provided in Appx.~\ref{sec:ablations}.
\section{Conclusion}
We demonstrated that domain-specific, multimodal text encoders, such as CXR-BERT, significantly enhance phrase grounding performance in LDMs, particularly in the medical imaging domain. By integrating such encoders, our approach nearly doubles key metrics like mIoU compared to state-of-the-art discriminative methods, establishing generative models as a superior alternative for this task. Additionally, we introduced BBM, which further refines cross-attention maps to improve localization accuracy and robustness.

Our findings highlight the untapped potential of generative models in aligning text and image modalities, providing a pathway toward more interpretable and trustworthy medical AI systems. While our work represents an advancement, it also underscores the importance of balancing interpretability with generative quality for clinical applications. Future research should focus on extending this approach to other medical domains and exploring strategies to optimize this balance further. 


\section{Acknowledgments}
\label{sec:acknowledgments}
The authors gratefully acknowledge the scientific support and HPC resources provided by the Erlangen National High
Performance Computing Center (NHR@FAU) of the Friedrich-Alexander-Universität Erlangen-Nürnberg (FAU) under the NHR
projects b143dc and b180dc.
NHR funding is provided by federal and Bavarian state authorities.
NHR@FAU hardware is partially funded by the German Research Foundation (DFG) – 440719683.1

\bibliography{midl25_036}

\appendix
\newpage
\section{Ablation study}
\label{sec:ablations}
\begin{table}[htbp]
\caption{Comparison of the average disease detection metrics across different sampling methods and fine-tuned
LLMs
using $timesteps=50$. The CLIP results are reproduced from~\citet{Dombrowski_2024}.}
\label{tab:llm_comparison}
\centering
\resizebox{\textwidth}{!}{%
\begin{tabular}{llccccc}
 \toprule
Sampling Input & Model & mIoU & Top-1 & AUC-ROC & CNR  \\
\midrule
\multirow{5}{*}{Text-Conditioning} & CLIP~\cite{Rombach_2022_CVPR} & 0.3540 & 0.1694 & 0.6304 & 0.3571 \\
& OpenCLIP~\cite{ilharco_2021_openclip} & 0.3337 & 0.0828 & 0.5676 & 0.1634 \\
& RadBERT~\cite{chambon_cook_langlotz_2022} & 0.3535 & 0.2929 & 0.6693 & 0.4901 \\
& Med-KEBERT~\cite{zhang_knowledge-enhanced_2023}& 0.3188 & 0.1400 & 0.5715 & 0.2011 \\
& CXR-CLIP~\cite{you_cxr-clip} & 0.3556 & 0.2477 & 0.6504 & 0.4319 \\
& CXR-BERT~\cite{Boecking_2022_MS_CXR} & 0.2932 & 0.0694 & 0.5895 & 0.2343 \\
\midrule
\multirow{5}{*}{CFG} & CLIP~\cite{Rombach_2022_CVPR} & 0.3838 & 0.2422 & 0.6606 & 0.4403 \\
& OpenCLIP~\cite{ilharco_2021_openclip} & 0.3226 & 0.0443 & 0.5512 & 0.1195 \\
& RadBERT~\cite{chambon_cook_langlotz_2022} & 0.3542 & 0.2709 & 0.6604 & 0.4558 \\
& Med-KEBERT~\cite{zhang_knowledge-enhanced_2023} & 0.3190 & 0.2134 & 0.5880 & 0.2496 \\
& CXR-CLIP~\cite{you_cxr-clip} & 0.3556 & 0.2477 & 0.6504 & 0.4319 \\
& CXR-BERT~\cite{Boecking_2022_MS_CXR} & 0.4541 & 0.4230 & 0.7530 & 0.7730 \\
\midrule
\multirow{5}{*}{GT-1 + CFG } & CLIP~\cite{Rombach_2022_CVPR} & 0.3528 & 0.2357 & 0.6607 & 0.4568 \\
& OpenCLIP~\cite{ilharco_2021_openclip}  & 0.3302 & 0.0435 & 0.5873 & 0.2245 \\
& RadBERT~\cite{chambon_cook_langlotz_2022} & 0.3162 & 0.2983 & 0.6894 & 0.5644 \\
& Med-KEBERT~\cite{zhang_knowledge-enhanced_2023} & 0.3336 & 0.2900 & 0.6367 & 0.3897 \\
& CXR-CLIP~\cite{you_cxr-clip} & 0.3588 & 0.2589 & 0.6669 & 0.4810   \\
& CXR-BERT~\cite{Boecking_2022_MS_CXR} & 0.4067 & 0.39823 & 0.8060 & 0.9886\\
\midrule
\multirow{5}{*}{GT + CFG} & CLIP~\cite{Rombach_2022_CVPR} & 0.4089 & 0.4062 & 0.7413 & 0.7164 \\
& OpenCLIP~\cite{ilharco_2021_openclip} & 0.3344 & 0.0534 & 0.6451 & 0.4041 \\
& RadBERT~\cite{chambon_cook_langlotz_2022}  & 0.3995 & 0.4138 & 0.7781 & 0.8363 \\
& Med-KEBERT~\cite{zhang_knowledge-enhanced_2023} & 0.3452 & 0.3508 & 0.6893 & 0.5536 \\
& CXR-CLIP~\cite{you_cxr-clip} & 0.3684 & 0.2695 & 0.6979 & 0.5809 \\
& CXR-BERT~\cite{Boecking_2022_MS_CXR}  & 0.5242 & 0.6506 & 0.8657 & 1.2288 \\
\bottomrule
\end{tabular}
}
\end{table}

\noindent
The ablation studies investigate the impact of different domain-specific text encoders on phrase grounding.
As shown in~\tableref{tab:llm_comparison}, four different sampling methods are compared: only using generic
text conditioning during sampling, applying Conditional Free Guidance (CFG) during sampling, additionally giving the noisy ground-truth
image in the first timestep (GT-1 + CFG)  and additionally giving the noisy ground-truth image in every step (GT + CFG).
The metrics AUC-ROC and Top-1~\cite{Dombrowski_2024} are also included here.
The most important detail that can be seen in~\tableref{tab:llm_comparison}, is that CXR-BERT performs by far the best
of all tested models.
There are several attributes that distinguish CXR-BERT from the rest that could be responsible for that.
Unlike the domain-agnostic CLIP models, CXR-BERT is trained on domain specific CXR data.
In contrast to RadBERT, CXR-BERT is trained in a multimodal manner.
Compared to Med-KEBERT, CXR-BERT does not rely on any report preprocessing.
In comparison to CXR-CLIP, CXR-BERT has a considerably more complex pretraining procedure.
Additionally, CXR-BERT uses both local and global loss, which differentiates it from all other discussed models.
Adding a local loss term, paired with the domain-specific, multimodal training is most likely the key to the strong
performance of CXR-BERT.
\section{Processing Methods Comparison}
\label{sec:processing}
\begin{table}[htbp]
\centering
\caption{Comparison of different processing methods on CXR-BERT\textsubscript{LDM} for our primary phrase grounding metrics. Disease filtering refers to the token processing method used by \citet{Dombrowski_2024}, while lexical filtering refers to our token processing method that filters words without lexical meaning. Linear Bézier refers to the linear interpolation of the image and text bias. The quadratic Bézier approach uses the multiplicative interaction of the biases as control point. Mixture Bézier blends the quadratic and linear approach.}
\label{tab:processing_comp}
\setlength{\tabcolsep}{6pt} 
\resizebox{\textwidth}{!}{%
\begin{tabular}{@{}lcccccccccc@{}}
    \toprule
    \multirow{2}{*}{Disease} & \multicolumn{2}{c}{Disease Filtering} & \multicolumn{2}{c}{Lexical Filtering} & \multicolumn{2}{c}{Linear Bézier} & \multicolumn{2}{c}{Quadratic Bézier} & \multicolumn{2}{c}{Mixture Bézier} \\
    \cmidrule(lr){2-3} \cmidrule(lr){4-5} \cmidrule(lr){6-7} \cmidrule(lr){8-9}  \cmidrule(lr){10-11}
    & mIoU & CNR & mIoU & CNR & mIoU & CNR & mIoU & CNR & mIoU & CNR
    \\
    \midrule

Atelectasis     & 0.506 & 1.53 & 0.589 & 1.71 & 0.548 & 1.79  & 0.576 & 1.78 & 0.559 & 1.79\\
Cardiomegaly    & 0.649 & 1.34 & 0.649 & 1.38 & 0.657 & 1.57  & 0.685 & 1.54 & 0.670 & 1.58\\
Consolidation   & 0.522 & 1.48 & 0.520 & 1.63 & 0.488 & 1.68  & 0.518 & 1.69 & 0.498 & 1.69\\
Lung Opacity    & 0.418 & 1.35 & 0.462 & 1.55 & 0.438 & 1.63  & 0.470 & 1.67 & 0.451 & 1.66\\
Edema           & 0.556 & 1.43 & 0.543 & 1.34 & 0.556 & 1.38  & 0.543 & 1.36 & 0.534 & 1.38\\
Pneumonia       & 0.510 & 1.48 & 0.536 & 1.59 & 0.508 & 1.70  & 0.539 & 1.68 & 0.516 & 1.70\\
Pneumothorax    & 0.308 & 0.71 & 0.336 & 0.55 & 0.318 & 0.41  & 0.356 & 0.36  & 0.332 & 0.37\\
Pl. Effusion    & 0.565 & 1.69 & 0.551 & 1.69 & 0.565 & 1.69  & 0.535 & 1.67 & 0.514 & 1.69\\
\midrule
Weighted Avg.   & 0.524 & 1.23 & 0.537 & 1.24 & 0.526 & 1.31 & 0.558 & 1.28 & 0.539 & 1.31\\
\bottomrule
\end{tabular}
}
\end{table}
\noindent 
In~\tableref{tab:processing_comp}, we can see how different processing methods affect the phrase grounding performance of our model.
The first two columns are concerned with which tokens should be considered for the creation of the activation maps.
Neither start nor end tokens are considered in either approach.
\citet{Dombrowski_2024} only used the tokens corresponding to the disease if at least one is present.
Otherwise, if the disease is not mentioned in the report, all tokens are used.
Meanwhile, our approach filters any words with no lexical meaning, and then uses the remaining tokens.
The approach is motivated by the findings seen in~\figureref{fig:tokens}.
This change yields a small improvement in phrase grounding performance compared to the original method.

The remaining three columns of~\tableref{tab:processing_comp} are concerned with different interpolation techniques that can be used for BBM.
Linear Bézier refers to a usual linear interpolation, meaning the equation 
\begin{equation}
    \label{eq:linear_int}
    sP_\text{mult} + (1-s)P_\text{comb}
\end{equation}
is used.
While this method improves CNR considerably, it is not well-suited for mask generation, since the larger activation areas lead to masks that are too large.
To fix this issue, Quadratic Bézier incorporates a quadratic Bézier curve for interpolation, namely
\begin{equation}
    \label{eq:quadratic_int}
    2(1-s)s(P_\text{mult}\odot P_\text{comb}) + (1-s)^2P_\text{comb} + s^2 P_\text{mult}. 
\end{equation}
The control point of the interpolation is the Hadamard product of $P_\text{mult}$ and  $P_\text{comb}$.
This serves primarily two purposes: first, to catch more complex multiplicative interactions between the two matrices.
Second, the multiplication with $P_\text{mult}$ acts as a gating mechanism, which means using the multiplicative interaction as the midpoint hinders the interpolation from having overly large areas of activation.
Instead, the interpolation has a greater focus on the relevant areas of activation.
Using this approach, the mask generation can be successfully improved, but at the cost of a lower improvement of CNR.
Consequently, we blend the linear and quadratic approach to gain the benefits of both, as can be seen in the Mixture Bézier column.
The result is an essentially linear interpolation which midpoint is combined with $P_\text{mult} \odot P_\text{comb}$.
Since a linear Bézier curve can be expressed as a quadratic Bézier curve with 
\begin{equation}
    \label{eq:lin_as_quad_int}
    2(1-s)s\left(\frac{P_\text{mult} + P_{\text{comb}}}{2}\right)+ (1-s)^2 P_{\text{comb}} + s^2P_\text{mult},
\end{equation}
incorporating the Hadamard product results in
\begin{equation}
    \label{eq:mixture_int}
    2(1-s)s\left(\frac{P_\text{mult} + P_{\text{comb}} + P_\text{mult} \odot P_{\text{comb}}}{2}\right) 
+ (1-s)^2 P_{\text{comb}} + s^2P_{\text{mult}}.
\end{equation}
This interpolation achieves a balance between the accuracy of the activation maps and their corresponding masks.

\section{Hyperparameter Optimization}
\label{sec:timesteps}
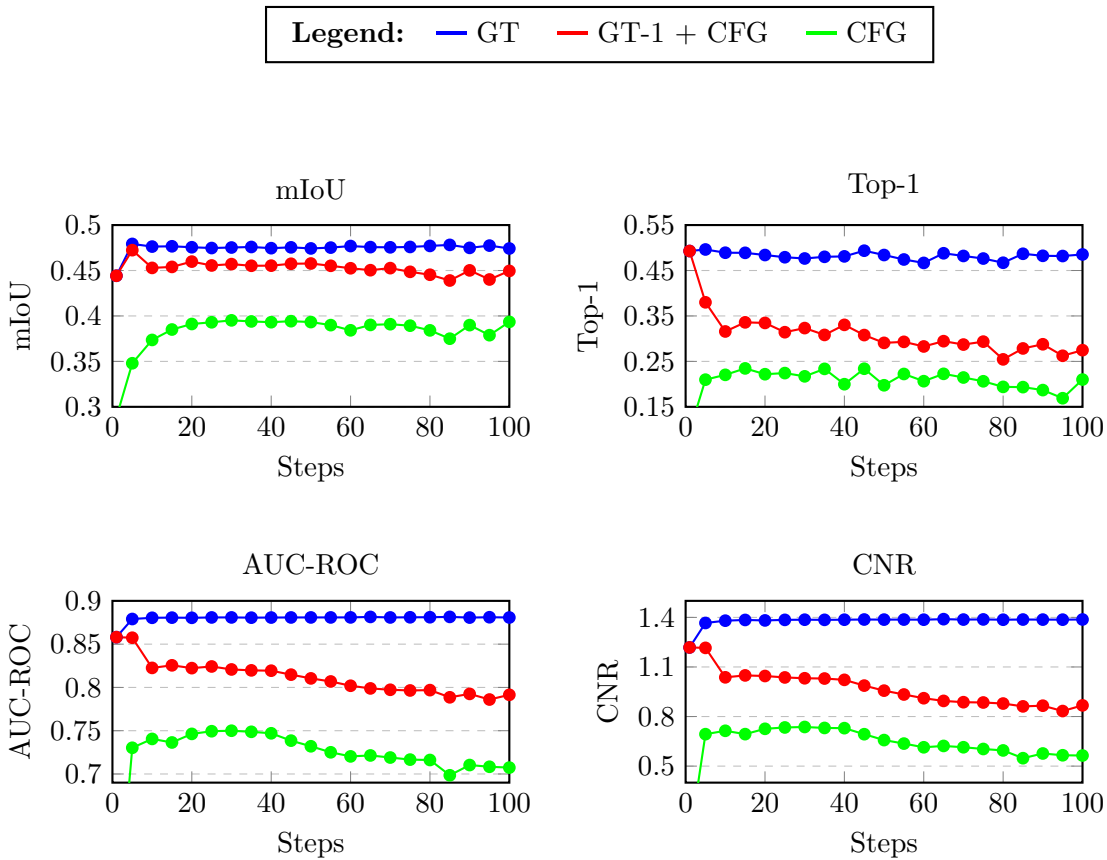
\begin{figure}[t]
\centering
\begin{tikzpicture}
    \begin{axis}[
        at={(0,0)},
        width=0.45\textwidth,
        height=4cm,
        xlabel={Steps},
        ylabel={mIoU},
        title={mIoU},
        xmin=0, xmax=100,
        ymin=0.3, ymax=0.5,
        xtick={0,20,40,60,80,100},
        ytick={0.3, 0.35, 0.4, 0.45, 0.50},
        ymajorgrids=true,
        grid style=dashed,
        thick,
    ]
    \addplot[color=blue, mark=*] table {
        1   0.4442203300044443
        5   0.4792295197005987
        10  0.4763143150579579
        15  0.4766305723222576
        20  0.4755837792132232
        25  0.474722005059809
        30  0.4753000647249974
        35  0.4758769468400256
        40  0.474566620069645
        45  0.4754716090581702
        50  0.4742933004062851
        55  0.4751925763466545
        60  0.4769044439454039
        65  0.4757250487933895
        70  0.4755111421892048
        75  0.4758941182060179
        80  0.4770387076703766
        85  0.4781768353134413
        90  0.4749073523638643
        95  0.4773905369290004
        100 0.4742229133931806
    };
    
    \addplot[color=red, mark=*] table {
        1   0.4442203300044443
        5   0.4723741753876913
        10  0.4528609908302062
        15  0.4539490163583879
        20  0.4597916666219048
        25  0.4555368111897871
        30  0.4570182733398955
        35  0.4552115009359674
        40  0.4553301861417677
        45  0.457416186549107
        50  0.4576959234187333
        55  0.455178842970234
        60  0.4524623384708238
        65  0.450337533594351
        70  0.4526109757983847
        75  0.4485443533117685
        80  0.4453418022191484
        85  0.4389928750527901
        90  0.4501450511847605
        95  0.440113269666649
        100 0.449542542483379
    };

   \addplot[color=green, mark=*]
   table{
        1   0.28850042732377523
        5   0.347943320459816
        10  0.3735984242032421
        15  0.3850650147254606
        20  0.3911445313677621
        25  0.3930282260327912
        30  0.3951535535291572
        35  0.393961133853564
        40  0.3931291188415233
        45  0.3942400878558822
        50  0.3932834991725266
        55  0.3899014667589002
        60  0.3841951826374594
        65  0.3901204077138983
        70  0.3909028195659551
        75  0.389293314591544
        80  0.3841373571015897
        85  0.3749756819958942
        90  0.389912246559251
        95  0.3788345412477052
        100 0.393478213828496

   };          
    \end{axis}

    \begin{axis}[
        at={(0.5\textwidth,0)},
        width=0.45\textwidth,
        height=4cm,
        xlabel={Steps},
        ylabel={Top-1},
        title={Top-1},
        xmin=0, xmax=100,
        ymin=0.15, ymax=0.55,
        xtick={0,20,40,60,80,100},
        ytick={0.15, 0.25, 0.35, 0.45, 0.55},
        ymajorgrids=true,
        grid style=dashed,
        thick,
    ]
    \addplot[color=blue, mark=*] table {
        1   0.4928617680614227
        5   0.4962612932071166
        10  0.4891307699423391
        15  0.4888840923807468
        20  0.4840661712558971
        25  0.4791942894144491
        30  0.4764500015417348
        35  0.4801193302704202
        40  0.4810752058215904
        45  0.4935786747248003
        50  0.4840738799296969
        55  0.4742376121612038
        60  0.466698529185039
        65  0.4879359255033764
        70  0.4820156640251611
        75  0.4763189540871388
        80  0.4670839628750269
        85  0.486825876476211
        90  0.4823163023033517
        95  0.4818691992229656
        100 0.4852918503900589
    };
    \addplot[color=red, mark=*] table {
        1   0.4928617680614227
        5   0.3799913662853442
        10  0.3158474915975455
        15  0.3359054608245198
        20  0.3346258209737596
        25  0.3142209614257963
        30  0.3232940704881132
        35  0.3082390305571829
        40  0.3304014677314915
        45  0.3079306836051925
        50  0.2907557583793284
        55  0.2929989824550584
        60  0.2828466590607751
        65  0.2944790478246122
        70  0.2870324689340446
        75  0.2934692115568437
        80  0.2544401961086615
        85  0.2784835496901113
        90  0.2876722888594246
        95  0.2623338780796151
        100 0.2746908821806296
    };

    \addplot[color=green, mark=*] table {
        1   0.09599611482840492
        5   0.2097992661342542
        10  0.2204064012827233
        15  0.2345672350528815
        20  0.2217785452190805
        25  0.224260738182603
        30  0.2171070888964262
        35  0.2335342727637137
        40  0.1998011162159662
        45  0.2338426197157041
        50  0.1974499707070395
        55  0.2222950263636643
        60  0.2064537017051586
        65  0.2226110819894545
        70  0.2145246831735068
        75  0.2062301501649656
        80  0.1938885634115506
        85  0.1933720822669667
        90  0.1868505442323702
        95  0.1689510036693287
        100 0.2100459436958465
    };
    \end{axis}

    \begin{axis}[
        at={(0,-5cm)},
        width=0.45\textwidth,
        height=4cm,
        xlabel={Steps},
        ylabel={AUC-ROC},
        title={AUC-ROC},
        xmin=0, xmax=100,
        ymin=0.69, ymax=0.9,
        xtick={0,20,40,60,80,100},
        ytick={0.7, 0.75, 0.8, 0.85, 0.9},
        ymajorgrids=true,
        grid style=dashed,
        thick,
    ]
    \addplot[color=blue, mark=*] table {
        1   0.8580735144392647
        5   0.8790447065300185
        10  0.8804594380665782
        15  0.8806242463923664
        20  0.8804331773878884
        25  0.8808505754467764
        30  0.880871752858659
        35  0.8805981096669194
        40  0.8807346993491364
        45  0.8808339949120385
        50  0.8808616757693208
        55  0.8808683149598998
        60  0.880903052361633
        65  0.8814353909895344
        70  0.8808469868889551
        75  0.8810246477470701
        80  0.881098143595408
        85  0.8814689946936798
        90  0.8806932085760294
        95  0.8811003762621725
        100 0.8807954756799814
    };
        \addplot[color=red, mark=*] table {
        1   0.8580735144392647
        5   0.8574078927255114
        10  0.8225522868588313
        15  0.8255310836208508
        20  0.82229425111488
        25  0.8242012347857575
        30  0.8207565708220131
        35  0.8197129912201349
        40  0.8193308002255981
        45  0.8147641326196631
        50  0.8104330100871762
        55  0.8068805153252403
        60  0.8019332462327624
        65  0.7988607717010868
        70  0.7972905762057038
        75  0.796491551246358
        80  0.796886021518248
        85  0.7885844771255736
        90  0.7925269264426883
        95  0.7860197158828971
        100 0.791470900495925
    };

    \addplot[color=green, mark=*]
    table {
        1   0.5203382992145659
        5   0.7302793420893162
        10  0.7405053359742517
        15  0.7364058549767514
        20  0.7463180276819446
        25  0.749407289449176
        30  0.749975963600224
        35  0.7487251865529068
        40  0.7470144726738933
        45  0.7385197955800858
        50  0.7320046921968817
        55  0.7250473479867713
        60  0.7203211897012592
        65  0.7214053812690505
        70  0.7190074816171164
        75  0.7166066948637537
        80  0.7161566676755068
        85  0.6984895400298874
        90  0.7103341785620358
        95  0.7083307955434153
        100 0.7073876598999287
    };
    \end{axis}

    \begin{axis}[
        at={(0.5\textwidth,-5cm)},
        width=0.45\textwidth,
        height=4cm,
        xlabel={Steps},
        ylabel={CNR},
        title={CNR},
        xmin=0, xmax=100,
        ymin=0.4, ymax=1.5,
        xtick={0,20,40,60,80,100},
        ytick={0.5, 0.8, 1.1,  1.4},
        ymajorgrids=true,
        grid style=dashed,
        thick,
    ]
    \addplot[color=blue, mark=*] table {
        1   1.2180771327374431
        5   1.3661480145692528
        10  1.3801359602349497
        15  1.3844940072060965
        20  1.3816801070644622
        25  1.384970131339044
        30  1.38602910540397
        35  1.3851629839993769
        40  1.3859916528045586
        45  1.3871194116681569
        50  1.3865686507868609
        55  1.3868023034830037
        60  1.3861102275095334
        65  1.3892431977917528
        70  1.3868203157919392
        75  1.3876297927992198
        80  1.3859711013974534
        85  1.3869756194568974
        90  1.3866413766269796
        95  1.3862841719256225
        100 1.3873784883988372
    };
    \addplot[color=red, mark=*] table {
        1   1.2180771327374431
        5   1.2155110808742615
        10  1.0373296115576645
        15  1.0488725659187486
        20  1.0453220656378368
        25  1.036626484518696
        30  1.0316104709762608
        35  1.0300479871727133
        40  1.0220295107461
        45  0.987223325759974
        50  0.9567183891540564
        55  0.933385664808163
        60  0.9108679703672892
        65  0.8945714896418783
        70  0.8868169004437713
        75  0.8849993173917624
        80  0.8786778536708651
        85  0.8614876371566581
        90  0.8656010490797135
        95  0.833216551034743
        100 0.8675572369918362
    };
    \addplot[color=green, mark=*]
    table {
        1   0.051766828908138426
        5   0.6932932632942425
        10  0.7137350349479401
        15  0.6935003220107385
        20  0.7257326405135739
        25  0.7339184627975118
        30  0.7368253992938282
        35  0.7309459694215436
        40  0.7297143413380136
        45  0.6930906198594983
        50  0.657561418303753
        55  0.6363062429500282
        60  0.6139472164047026
        65  0.622100520785896
        70  0.6141756784653954
        75  0.6036174734921275
        80  0.5948910493288434
        85  0.5479222859761153
        90  0.5764287934247495
        95  0.5656823312001966
        100 0.5641376385273237
    };
    \end{axis}
    
\path (0,0.35\textwidth) -- (0.85\textwidth,0.35\textwidth) node[midway,below,draw=black, thick] {
    \begin{tabular}{cc@{\hskip 0.5cm}c@{\hskip 0.5cm}c}
        \textbf{Legend:} &
        \textcolor{blue}{\textbf{\rule[0.5ex]{0.4cm}{1.5pt}}} GT &
        \textcolor{red}{\textbf{\rule[0.5ex]{0.4cm}{1.5pt}}} GT-1 + CFG &
        \textcolor{green}{\textbf{\rule[0.5ex]{0.4cm}{1.5pt}}} CFG
    \end{tabular}
};
\end{tikzpicture}
\caption{Performance of CXR-BERT using different sampling modes, trained on the MIMIC dataset, across various timesteps when evaluated on the ChestXRay14 dataset.
The shown sampling modes are CFG with the ground truth image in the initial denoising step (GT-1 + CFG)
    and CFG with the ground truth image in every denoising step (GT). }
\label{fig:timestep_comparison}
\end{figure}
All hyperparemter optimization was carried out on the ChestXRay14~\cite{wang_2017_chestxray} dataset.
In addition to CNR and mIoU, we also include the Top-1 metric~\cite{Dombrowski_2024} and the AUC-ROC metric here.
\figureref{fig:timestep_comparison} demonstrates that after about five timesteps, there are no significant improvements
in phrase grounding performance during image generation using a noisy ground-truth image as input.
This indicates that our results do not depend heavily on this hyperparameter.
It seems reasonable that the results saturate early for this sampling method compared to the others, since the model is given
significantly more information in form of the ground-truth images.
Using the noisy ground-truth image only in the first denoising step is the same to our main approach initially.
This behavior is intuitive, since using the ground-truth only in the first step means essentially using it in all steps if only a single timestep is used in total.
As the number of timesteps increases, the performance gradually becomes closer to the results from text-only conditioning.
However, while there is a steep decline after the first few steps if the ground-truth image is only used in the first
sampling step, the performance eventually stabilizes.

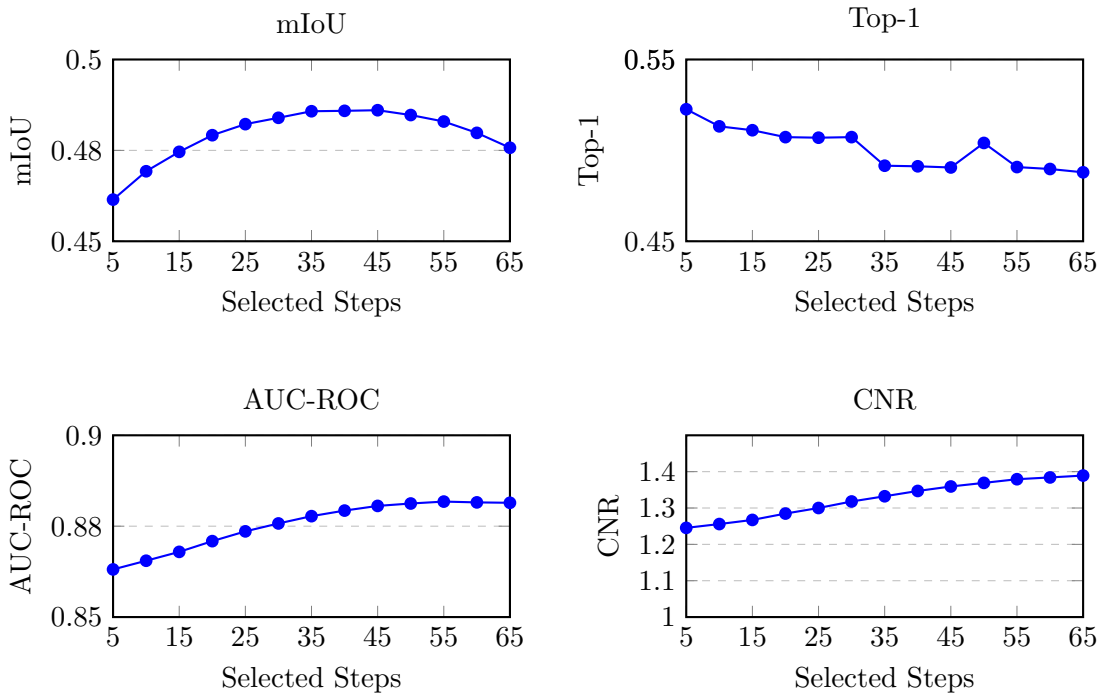
\begin{figure}[t]
\centering
\begin{tikzpicture}
    \begin{axis}[
        at={(0,0)},
        width=0.45\textwidth,
        height=4cm,
        xlabel={Selected Steps},
        ylabel={mIoU},
        title={mIoU},
        xmin=5, xmax=65,
        ymin=0.45, ymax=0.5,
        xtick={5,15,25,35,45,55,65},
        ytick={0.45, 0.475, 0.5},
        ymajorgrids=true,
        grid style=dashed,
        thick,
    ]
    \addplot[color=blue, mark=*] table {
        5	0.461460491
        10	0.469256354
        15	0.474598081
        20	0.479175842
        25	0.482216381
        30	0.483962012
        35	0.485766078
        40	0.485879847
        45	0.486050059
        50	0.484720585
        55	0.482939145
        60	0.47981683
        65	0.475725049

    };
    \end{axis}

    \begin{axis}[
        at={(0.5\textwidth,0)},
        width=0.45\textwidth,
        height=4cm,
        xlabel={Selected Steps},
        ylabel={Top-1},
        title={Top-1},
        xmin=5, xmax=65,
        ymin=0.45, ymax=0.55,
        xtick={5,15,25,35,45,55,65},
        ytick={0.45, 0.55, 0.55},
        ymajorgrids=true,
        grid style=dashed,
        thick,
    ]
    \addplot[color=blue, mark=*] table {
        5	0.522624958
        10	0.513220376
        15	0.511085073
        20	0.507284697
        25	0.50697635
        30	0.507292405
        35	0.491566711
        40	0.491242947
        45	0.490595418
        50	0.504085597
        55	0.49081897
        60	0.489755173
        65	0.487935926

    };
    \end{axis}

    \begin{axis}[
        at={(0,-5cm)},
        width=0.45\textwidth,
        height=4cm,
        xlabel={Selected Steps},
        ylabel={AUC-ROC},
        title={AUC-ROC},
        xmin=5, xmax=65,
        ymin=0.85, ymax=0.9,
        xtick={5,15,25,35,45,55,65},
        ytick={0.85, 0.875, 0.9},
        ymajorgrids=true,
        grid style=dashed,
        thick,
    ]
    \addplot[color=blue, mark=*] table {
        5	0.863106307
        10	0.865497599
        15	0.867934089
        20	0.870887545
        25	0.873574053
        30	0.875751882
        35	0.877757988
        40	0.879291617
        45	0.880577596
        50	0.88123863
        55	0.881766306
        60	0.881545827
        65	0.881435391

    };
    \end{axis}

    \begin{axis}[
        at={(0.5\textwidth,-5cm)},
        width=0.45\textwidth,
        height=4cm,
        xlabel={Selected Steps},
        ylabel={CNR},
        title={CNR},
        xmin=5, xmax=65,
        ymin=1.0, ymax=1.5,
        xtick={5,15,25,35,45,55,65},
        ytick={1.0, 1.1, 1.2, 1.3, 1.4},
        ymajorgrids=true,
        grid style=dashed,
        thick,
    ]
    \addplot[color=blue, mark=*] table {
        5	1.245379985
        10	1.255878153
        15	1.267221734
        20	1.284713352
        25	1.300015977
        30	1.317945212
        35	1.332299718
        40	1.346891249
        45	1.359206707
        50	1.369126178
        55	1.379205036
        60	1.384091564
        65	1.389243198

    };
    \end{axis}
\end{tikzpicture}
\caption{Results for LDM ground-truth sampling conditioned by CXR-BERT on the
MIMIC-CXR dataset.}
\label{fig:timestep_selection}
\end{figure}

Figure~\ref{fig:timestep_selection} showcases how the selection of the last $n$ timesteps during
ground-truth sampling affects the results.
The figure is constrained to 65 timesteps, since this configuration produced the best results given all of the four phrase
grounding metrics.
Also, this can be seen as an exemplary result, since the observed trends are very similar for all timesteps in a sensible
range.
As can be seen in~\figureref{fig:timestep_selection}, the metrics do not show the same behavior over the number of selected timesteps.
CNR and AUC-ROC steadily increase the more timesteps are selected.
However, the progression of mIoU is concave, peaking at selecting 45 timesteps and then decreasing.
Meanwhile, Top-1 peaks at 5 timesteps and then shows a tendency to increase over time.

Since AUC-ROC and CNR are closely related metrics, it makes sense that they again show similar behavior.
When incorporating more timesteps, noise introduced in single timesteps becomes less relevant.
Both AUC-ROC and CNR give worse results when more noise is introduced to the signal, which is why these metrics typically perform better for a larger number of timesteps.
Meanwhile, Top-1 only incorporates the highest activations, which is why this metric is extremely robust to noise. 
Therefore, only selecting a low number of timesteps can work well.
Top-1 most likely decreases when selecting a larger number of timesteps, since early timesteps focus on coarse features, resulting in larger activation areas.
Consequently, it is more likely that the highest activation is no longer strictly within the ground-truth bounding box.
Meanwhile, the mIoU metric can tolerate a certain amount of noise, due to the thresholding applied when creating the binary masks.
However, our masking approach generally has a tendency to include too much of the signal as part of the mask.
So both too much noise and too large activation areas decrease the quality of the generated mask.
Therefore, a compromise between both, resulting in a selection of roughly half of the timesteps, results in the best mIoU values in this setup.

When looking at the results, one should keep in mind that the changes are all very low, so the number of selected
timesteps does not play a considerable role as a hyperparameter either.
\section{Interpretability Trade-off}
\label{sec:tradeoff}
\begin{table}[htbp]
    \centering
    \caption{Comparison of FID and FID\textsubscript{XRV} values for images generated by LDMs using different text encoders.
        CNR values are also included to highlight the negative correlation between CNR and FID. The frozen CLIP
        results is reproduced and the learnable CLIP result is adopted from~\citet{Dombrowski_2024}.}
    \label{tab:fid_comparison}
    \begin{tabular}{l@{\hskip 0.3cm}c@{\hskip 0.5cm}c@{\hskip 0.5cm}c}
        \toprule
        Model & FID & FID\textsubscript{XRV} & CNR \\
        \midrule
        Frozen CXR-BERT & 109.4 & 18.0 & 0.84 \\
        Frozen CLIP & 83.8 & 14.5 & 0.72\\
        Learnable CLIP & 61.9 & 7.7 & 0.13 \\
        \bottomrule
    \end{tabular}
\end{table}
While the presented phrase grounding performance of our model allows for a higher degree of interpretability than comparable models, this comes at a price.
\citet{Dombrowski_2024} discuss a trade-off between interpretability and performance in LDMs.
They note that models with weaker phrase grounding capabilities often produce lower-quality images, while those with
higher image quality tend to have poorer phrase grounding performance.
\tableref{tab:fid_comparison} shows that this pattern is evident in our experiments as well.
Using a frozen CXR-BERT encoder for text conditioning results in lower FID scores, but produces the best
phrase grounding performance.
Meanwhile, the frozen CLIP encoder, which has weaker phrase grounding, achieves better FID
scores.
A learnable CLIP encoder provides the highest image quality, but the lowest phrase grounding metrics.
This means that choosing the correct model for application in the clinical field needs to be carefully considered. 
Simply using the model with the best image generation capabilities might produce good images.
However, no trust can be put into the fidelity of these images, since their internal representations cannot be interpreted.
Additionally, their lacking alignment between the image and text modalities imply that these models have no proper understanding of what they are generating, which might lead to harmful biases and mistakes in the generated images.
Models with better phrase grounding capabilities might be more trustworthy, but their generated images lacking in quality can also be problematic for clinical applications.
Even if the model has a good internal representation of the modalities, if the model generates subpar or unrealistic images, these can hardly be used in clinical settings.
Currently, professionals need to choose a fitting balance between between quality and interpretability depending on their use case.
\newpage
\section{Attention Maps on Different Resolutions}
\label{sec:layers}
\begin{figure}[ht]
  \centering
  \fbox{\textbf{LoRA-CLIP}}\\[0.5ex]
  \subfigure[All layers]{%
    \includegraphics[width=0.3\textwidth]{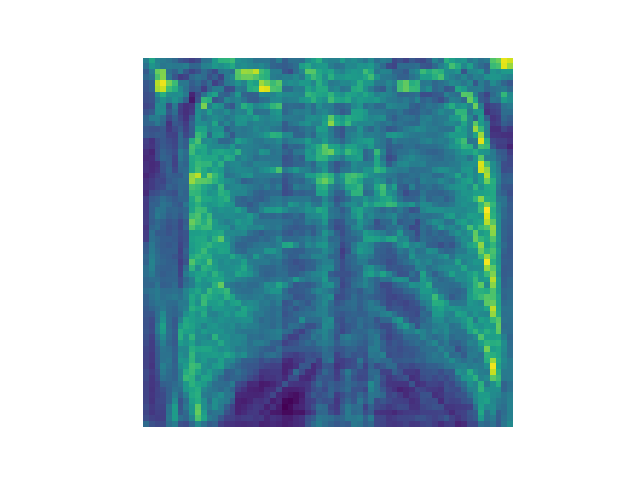}
    \label{fig:clip_all}
  }
  \quad
  \subfigure[Layers 5--7]{%
    \includegraphics[width=0.3\textwidth]{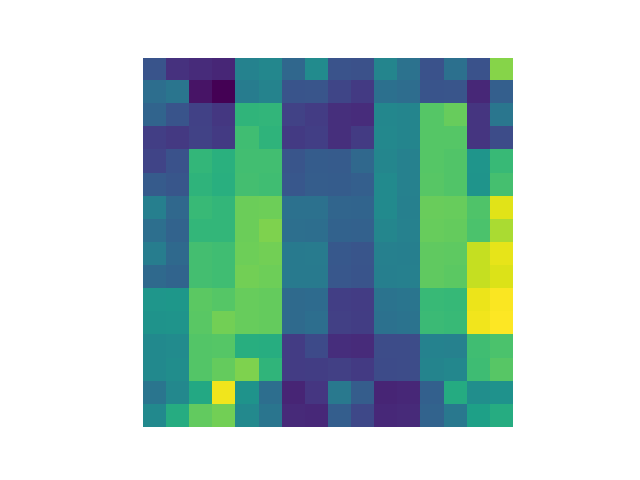}
    \label{fig:lora_layers}
  }
  
  \vspace{1ex}
  \hrule
  \vspace{1ex}
  
  \fbox{\textbf{CXR-BERT}}\\[0.5ex]
  \subfigure[All Layers]{%
    \includegraphics[width=0.3\textwidth]{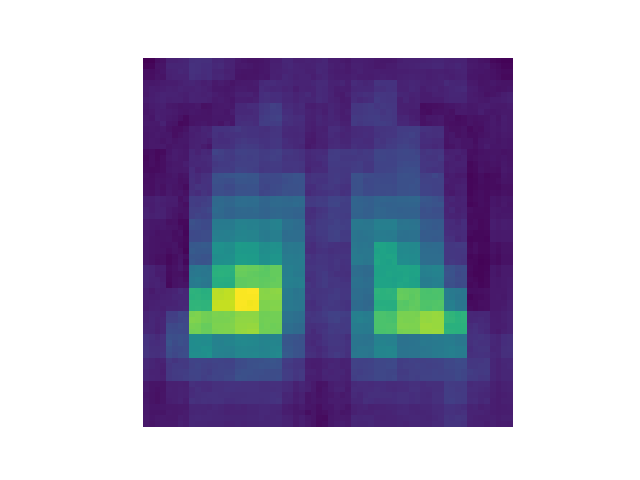}
    \label{fig:cxrbert_all}
  }
  \quad
  \subfigure[Layers 5--7]{%
    \includegraphics[width=0.3\textwidth]{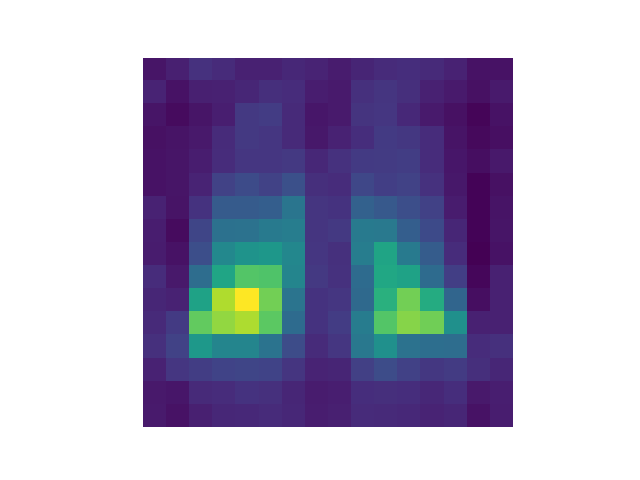}
    \label{fig:cxrbert_layers}
  }
  
  \caption{Comparison of cross-attention maps between conditioning provided by a CLIP model fine-tuned via LoRA~\cite{hu2022lora} and one conditioned by CXR-BERT. It can be seen that LoRA provides no localization capabilities and can consequently profit from the lower resolution through selecting the middle layers, while this does not work on CXR-BERT.}
  \label{fig:layers}
\end{figure}

As can be seen in \figureref{fig:layers}, models that have not learned a proper alignment between the image and text modalities can benefit from only selecting the innermost attention layers (\textit{i.e.}, the layers with the lowest resolution. 
However, this is mainly due to lower resolutions naturally being closer to an activation area compared to fine-grained features.
Inspecting all layers highlights that, in truth, such models have not learned a proper alignment between the two modalities and instead focus on unnecessary details such as the ribcage.
In contrast, CXR-BERT has learned a proper alignment, so we can simply average over all layers to get our results.

\end{document}